\pdfoutput=1

\documentclass[11pt]{article}

\usepackage[]{acl}

\usepackage{times}
\usepackage{latexsym}

\usepackage[T1]{fontenc}

\usepackage[utf8]{inputenc}

\usepackage{microtype}

\usepackage{inconsolata}

\usepackage{graphicx}

\newcommand\blfootnote[1]{
    \begingroup
    \renewcommand\thefootnote{}\footnote{#1}
    \addtocounter{footnote}{-1}
    \endgroup
}

\title{RETUYT-INCO at BEA 2025 Shared Task:\\ How Far Can Lightweight Models Go in AI-powered Tutor Evaluation?}

\author{Santiago Góngora $\dagger$ \and Ignacio Sastre $\dagger$ \and Santiago Robaina \\ {\bf Ignacio Remersaro} \and {\bf Luis Chiruzzo} \and {\bf Aiala Rosá}\\\\
Instituto de Computación, Facultad de Ingeniería, Universidad de la República \\ Montevideo, Uruguay}

\begin{document}
\maketitle
\begin{abstract}
In this paper, we present the RETUYT-INCO participation at the BEA 2025 shared task.
Our participation was characterized by the decision of using relatively small models, with fewer than 1B parameters.
This self-imposed restriction tries to represent the conditions in which many research labs or institutions are in the Global South, where computational power is not easily accessible due to its prohibitive cost.
Even under this restrictive self-imposed setting, our models managed to stay competitive with the rest of teams that participated in the shared task.
According to the $exact\ F_1$ scores published by the organizers, the performance gaps between our models and the winners were as follows: $6.46$ in \textit{Track 1}; $10.24$ in \textit{Track 2}; $7.85$ in \textit{Track 3}; $9.56$ in \textit{Track 4}; and $13.13$ in \textit{Track 5}.
Considering that the minimum difference with a winner team is $6.46$ points --- and the maximum difference is $13.13$ --- according to the $exact\ F_1$ score, we find that models with a size smaller than 1B parameters are competitive for these tasks, all of which can be run on computers with a low-budget GPU or even without a GPU.
\end{abstract}

\section{Introduction}
\label{sec:introduction}

\blfootnote{$\dagger$ These (corresponding) authors contributed equally to this work: \texttt{\{sgongora,isastre\}@fing.edu.uy}}

The remarkable advances in the development of Large Language Models (LLMs) in recent years have turned Natural Language Processing into a discipline with great potential for application in different domains, and \textit{Education} is not the exception~\citep{ignat-etal-2024-solved}. 
However, these technological advances are not affordable to everyone. 
The cost of closed models --- which are the most powerful and are typically considered the State-of-the-art in NLP --- and the expensive infrastructure required to use large open models, coupled with negative effects on the environment, make research on other methods still essential.

Our RETUYT-INCO team, as a research lab from South America, is no exception to this reality.
Naturally, we are concerned about these issues and, consequently, we have focused on experimenting with open models in recent editions of the BEA shared tasks. 
For the 2023 shared task, consisting in generating teacher responses in educational dialogues~\citep{tack-etal-2023-bea}, we participated using open models, obtaining competitive results~\citep{baladon-etal-2023-retuyt}. One of the highlights of our participation was the \textit{``Hello'' baseline}, a simple strategy we followed which achieved remarkable results, unveiling the fragility of BERTScore~\citep{Zhang*2020BERTScore}. 
More recently, for the 2024 BEA shared task, consisting in performing simplification experiments for different languages~\citep{shardlow-etal-2024-bea}, we mainly focused on fine-tuning BERT and Mistral models (i.e., open models), even using synthetic data in some cases~\citep{sastre-etal-2024-retuyt}.

\begin{figure*}
    \centering
    \includegraphics[trim={0.5cm 1.2cm 0 1.2cm},clip, width=\linewidth]{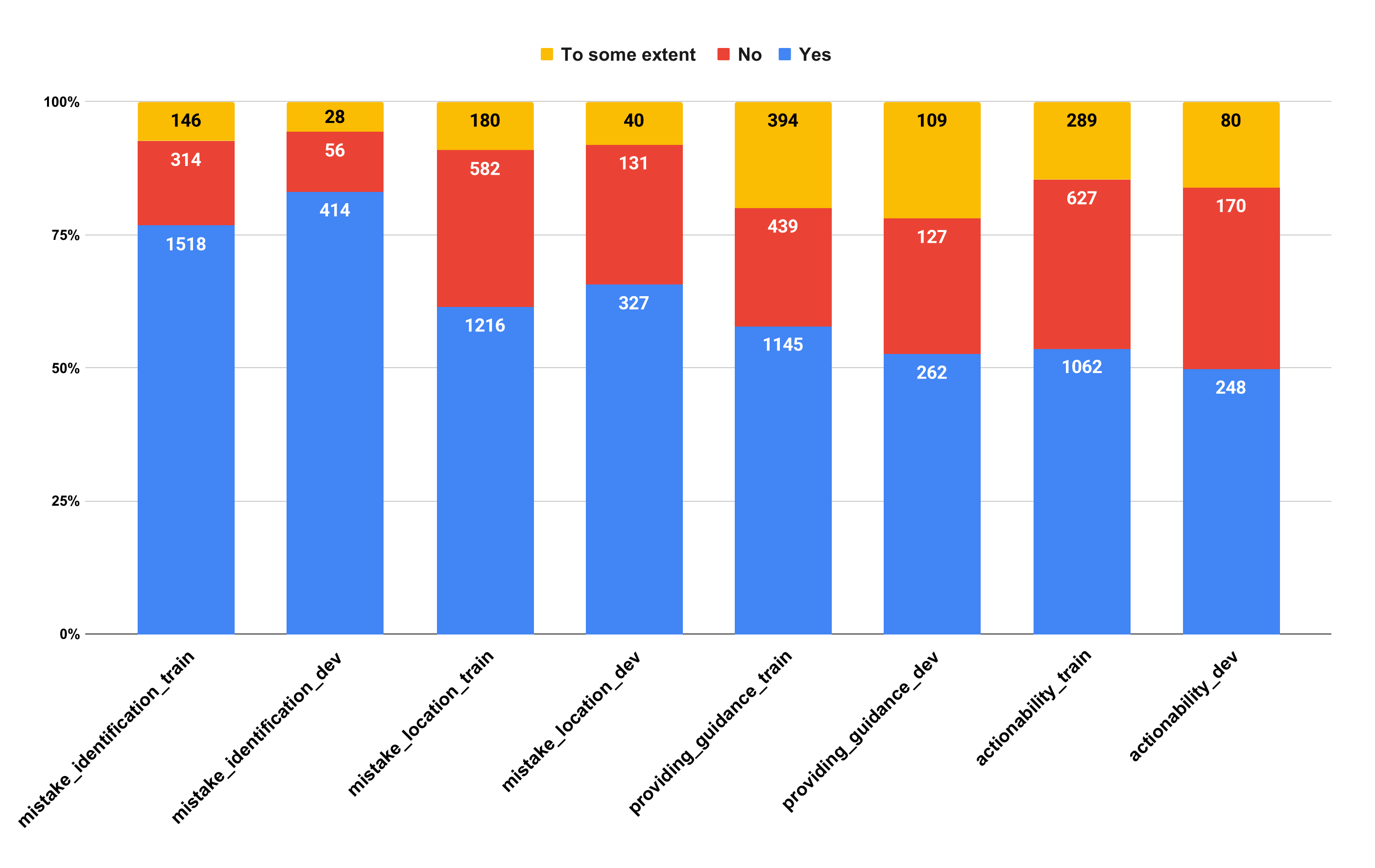}
    \caption{Class balance in our \texttt{train} and \texttt{dev} sets. The columns are coupled according to classes.}
    \label{fig:chart_train_dev_balance}
\end{figure*}

In this paper, we present the RETUYT-INCO participation in the\textbf{ five tracks} of the BEA 2025 Shared Task: \textit{Pedagogical Ability Assessment of AI-powered Tutors}~\cite{kochmar2025bea}.
This year, in addition to maintaining our restriction of working with open models, we challenged ourselves with an extra restriction: to experiment only with language models of fewer than a billion parameters and classical machine learning (ML) approaches.
We will call these \textit{lightweight} models, as they have to be small enough to run on a low-end GPU or with no GPU at all. 
This restriction is related to the situation many research labs face every day in the Global South: the lack of minimum resources to run what other regions consider \textit{small} models (7B parameters or more).
In our case, we have limited access to a national computing cluster, which we can use to fine-tune LLMs up to 7B parameters, but we do not have resources to host the fine-tuned LLMs and use them in real applications.

Moreover, this is not the only motivation for this self-imposed restriction, as one of the research lines of our lab is the application of NLP tools to aid teachers in rural areas
\citep{chiruzzo-etal-2022-using, rosá2025platformgeneratingeducationalactivities}.
In such contexts it is very unlikely to use state-of-the-art LLMs, due to the impossibility of using them through APIs (since children's privacy is key, sending private data to third-party servers is not an option), and the prohibitive cost of installing capable GPUs in trustworthy servers.

Overall, throughout this paper, we will try to answer the general research question that motivated our participation:
\textit{What is the performance gap between lightweight models and those state-of-the-art models, which would naturally have a better chance of winning the competition?}.

\section{Dataset}
\label{sec:dataset}

For this edition, the dataset consists of 300 conversations~\cite{maurya-etal-2025-unifying}.
Each dialogue is composed of interactions between a teacher and a math student.
In the final turn of each dialogue the student shows clear confusion about a concept, and the dataset includes potential tutor \textit{responses} intended to help the student.
These responses --- some of them generated by seven LLM-based tutors and others written by human tutors ---  are also evaluated by human evaluators (using \textbf{Yes}, \textbf{No} or \textbf{To some extent}) according to four dimensions of interest that coincide with the four proposed tracks in the shared-task: \textit{mistake identification}, \textit{mistake location}, \textit{providing guidance} and \textit{actionability}.

Due to the lack of a specific \textit{development} set, during the first month we split the official dataset published by the organizers into two parts: 80\%
We decided to do this split focusing on the conversations --- and not on the responses --- trying to ensure that each conversation and all its responses stayed either in the \texttt{train} or the \texttt{dev} set. 
As a consequence, our \textit{train-dev} split may not preserve the class balance of the original set.
Figure~\ref{fig:chart_train_dev_balance} shows the class balance for each dimension in our \texttt{train} and \texttt{dev} partition.

\section{Considered approaches}
\label{sec:approaches}

For our experiments we considered classical ML classification algorithms, BERT-based approaches and fine-tuning a small autoregressive language model.
Since all the tracks in the shared task are classification problems, many of the models we considered were used in more than one track.
All of them were trained (or fine-tuned) using our \texttt{train} set, running on GoogleColab\footnote{https://colab.research.google.com/} or a national computational cluster (see Section~\ref{sec:training_qwen}). 
At the end of this section --- in Subsection \ref{sec:results_dev} --- we will show the results we obtained when evaluating them on the \texttt{dev} set.

\subsection{Preliminary experiments}

To gain a greater understanding of how challenging the \textit{tracks} were, we performed four preliminary experiments with increasing degree of complexity.

The first and most basic one consisted in answering always \textbf{yes}, what naturally degraded the $F_1\ macro$ score, since the \textbf{No} and \textbf{To some extent} classes were never chosen.
Then, we tried using a random classifier, which consistently yielded accuracy values around 33\%

Additionally, before imposing ourselves the constraint of using \textit{lightweight} models only, we wanted to have an informed perception of how well bigger LLMs could perform in these tracks.
Therefore, we explored both prompting a closed model via an API and fine-tuning an open model.
For prompting, the model we chose was \texttt{Gemini Flash 2.0 Lite}\footnote{\url{https://deepmind.google/technologies/gemini/flash-lite/}}, using the prompt reported in the paper that presented the dataset~\citep{maurya-etal-2025-unifying}.
For fine-tuning, we chose \texttt{Llama 3.1 8B Instruct}~\cite{llama3modelcard}, and used Low Rank Adaptation (LoRA)~\cite{hu2022lora}.
This experiment follows the same methodology explained in Section \ref{sec:ftllm}.

\subsection{Classical Machine Learning approaches}
Among all the available algorithms in Sklearn\footnote{https://scikit-learn.org/stable/supervised\_learning.html} we tried, those that had the best performance on the \texttt{dev} set were \textit{Random Forest}, \textit{SVC} (Support-vector classifier) and \textit{k-NN} (k-Nearest Neighbors).
To represent the input texts we experimented with Bag of Words and TF-IDF, trying different n-gram ranges from $n=1$ to $n=8$. 

Since all these algorithms have problems capturing the complexities of long-context 
dependencies, after some experimentation we decided to train the models using the \textit{response} text only, i.e. without taking into consideration the full interaction between the student and the tutor.
Those preliminary experiments we did with the full interaction (i.e. concatenating the response of the tutor to the conversation history) had a notably lower performance.

\subsection{BERT-based approaches}

We also tried BERT-based approaches.
We experimented with fine-tuning them, combining them with classification algorithms and also with some rules. 

\subsubsection{BERT for Tracks 1--4}

We implemented a simple method by fine-tuning a simple BERT model for tracks 1 through 4. 
Specifically, we finetuned the DistilBERT \texttt{distilbert-base-uncased} variant~\cite{sanh2020distilbert}, a compact and computationally efficient distillation of BERT with approximately 66 million parameters.

For this experiment, we only considered the response text as input data, without the conversation history. We fine-tuned the model to predict each of the target variables (\texttt{mistake\_identification}, \texttt{mistake\_location}, \texttt{providing\_guidance}, \texttt{actionability}).
We initially tried to fine-tune the model in a three-class configuration, but our experiments were unable to predict any value of the class \textbf{To some extent} whatsoever, so we changed the approach.
We ended up training two-class models, joining \textbf{No} and \textbf{To some extent} as the negative class.
After fine-tuning, we analyzed the logit of the positive class and observed that even if both classes were lumped together during training, the \textbf{No} values actually got lower logit than the \textbf{To some extent} values, which allowed us to define thresholds to separate the three classes.

The hyperparameters in these experiments were the number of training epochs (from 1 to 3) and two thresholds to distinguish the frontier between \textbf{No} and \textbf{To some extent}, and between \textbf{To some extent} and \textbf{Yes}, which depending on the target output could vary between -1 and +1. In this round of experiments, we used Adam optimization with a learning rate of $5 \times 10^{-6}$.

\subsubsection{BERT for Track 5}

In our approach to track 5, the objective was to classify the tutor identity based once again solely on the provided response text. For fine-tuning, the following parameters were used: a learning rate of $2 \times 10^{-5}$, a weight decay of $0.01$, a training duration of $4$ epochs, and batch sizes set to $16$.

\subsubsection{Sentence Embeddings}

In addition to fine-tuning, we explored the use of BERT-like models to generate sentence embeddings~\cite{reimers-gurevych-2019-sentence}, which were then combined with classical ML methods for classification.

For tracks 1--4, we used the \texttt{multilingual-e5-large-instruct}\footnote{\url{https://huggingface.co/intfloat/multilingual-e5-large-instruct}} model \cite{wang2024multilingual}, a multilingual encoder initialized from \texttt{xlm-roberta-large}~\cite{conneau2019xlmroberta} (561M parameters).
We generated a sentence embedding for each example in our training partition and then used those embeddings as input to classical classifiers: k-NN and multilayer perceptron (MLP).

For this approach we explored three different input configurations:
\begin{itemize}
    \setlength\itemsep{0em}
    \item\textbf{Response-only:} The input consists solely of the embedding corresponding to the response to be evaluated.
    \item\textbf{Response + conversation history:} The input is formed by concatenating the embedding of the response with the embedding of the full conversation history.
    \item\textbf{Response + conversation history + LLM probabilities:} The input extends the previous configuration by appending the probabilities assigned to the three class labels by the fine-tuned LLM (see Section~\ref{sec:ftllm}).
\end{itemize}

For the k-NN classifier, we chose $k=9$ based on the performance on our \texttt{dev} set prior to submitting predictions for the competition's test set.
For the MLP, we used a simple model with no hidden layer and trained it until convergence, defined as no improvement greater than a tolerance of $1 \times 10^{-4}$ for 10 consecutive iterations.

For the mistake identification dimension, due to the high class imbalance in the training data, we applied under-sampling by fitting the k-NN classifier on a perfectly balanced subset.
This subset contained an equal number of examples for each class, matching the count of the least frequent class.
As shown in Section~\ref{sec:analysis}, this strategy led to improved performance.
For this classifier, different values of $k$ for each track were chosen (mistake identification: 415; mistake location: 540; providing guidance: 125; actionability: 96).

For track 5, we explored leveraging the DistilBERT model that was previously fine-tuned for direct sequence classification (as described in the previous section). In this setup, the core transformer layers (the base model, without the classification head) of this fine-tuned DistilBERT were employed as a feature extractor. Embeddings were generated for the "response" texts. These DistilBERT-derived embeddings were used as input features for an XGBoost classifier~\cite{chen2016xgboost}, which was configured for multiclass classification corresponding to the number of tutor labels. 

\subsubsection{BERT approach + Educated guess}
Another experimental approach we tried for track 5 was to take the predictions of the BERT + XGBoost model and, based on the distribution of the predicted labels, guess some tutor identities. Under the premise that if the model predicted correctly the majority of the time the correct tutor for a certain label, then taking the same prediction for a label might improve the performance of the model. 
Therefore, for this approach we modified the predictions of BERT + XGBoost forcing to always classify Tutor9 as ``Novice'', Tutor2 as ``Mistral'' and Tutor3 as ``Llama31405B''.

Unfortunately, this approach turned out to perform poorly in comparison to the BERT + XGBoost one, denoting that the labels shown in the test dataset might not have a direct mapping with the actual classes.

\subsection{Fine-tuning autoregressive LM}
\label{sec:ftllm}

In addition to using encoder-only transformers such as BERT, we also experimented with decoder-only LMs.
For these experiments, we only focused on the first four tracks.
Although these tracks are framed as classification and are therefore usually better suited to encoder-only architectures, we wanted to compare BERT-style fine-tuned models with similarly sized, fine-tuned autoregressive LMs.
Specifically, we used \texttt{Qwen2.5-0.5B-Instruct}\footnote{\url{https://huggingface.co/Qwen/Qwen2.5-0.5B-Instruct}}~\cite{qwen2.5,qwen2}, which has 494 million parameters and has undergone instruction tuning.

\subsubsection{Training}
\label{sec:training_qwen}

We performed full fine-tuning on our train partition.
Each example was converted into a prompt following the prompt template adopted during the model’s instruction tuning phase.
The prompt (available in Appendix~\ref{sec:appendix_prompts}) consists of:

\begin{itemize}
        \setlength\itemsep{0em}
	\item \textbf{System prompt:} We used the same system prompt reported in the shared-task dataset paper~\cite{maurya-etal-2025-unifying}, which was also used for LLM-based evaluation.
	\item \textbf{User message:} This part contains the conversation history, a task-specific rubric, and the response to be evaluated.
    The rubrics are the same as those used in the dataset paper.
	\item \textbf{Assistant message:} This consists solely of the class label corresponding to the example (\textbf{Yes} / \textbf{No} / \textbf{To some extent}).
\end{itemize}

We experimented with two different training approaches:

\begin{itemize}
    \setlength\itemsep{0em}

	\item \textbf{Dimension-specific approach:} This involves training four separate models, each dedicated to one of the four evaluation dimensions (mistake identification, mistake location, providing guidance, and actionability).
    Each model is trained only on examples corresponding to its specific dimension.
	\item \textbf{Multi-dimension approach:} This involves training a single model using the combined training data from all four dimensions.
    The model is expected to infer the appropriate evaluation criteria based on the scoring rubric included in the user message.
\end{itemize}

The multi-dimension approach may help mitigate the class imbalance present in certain dimensions (particularly mistake identification), as the model is exposed to a more balanced distribution of the three class labels across different contexts.

The model was trained for three epochs with a batch size of 8 and a learning rate set to $2 \times 10^{-4}$, using a linear scheduler with a warm-up ratio of 0.03 and weight decay of 0.001.
The training objective was next-token prediction, the same as in pre-training.

To train these models, we used the ClusterUY infrastructure~\cite{clusteruy} with limited (and usually interrupted) access to NVIDIA A100 and NVIDIA A40 GPUs.

\subsubsection{Inference}

Once the models are fine-tuned, we perform inference using greedy decoding.
The input prompt includes the system prompt and the user message, and the model is tasked with generating the assistant message.
Since these are classification tasks, we perform a single forward pass and select the class label whose first token receives the highest logit value.
Only the three candidate tokens (corresponding to the possible class labels) are considered, and the rest are ignored.
This approach prevents hallucinations by constraining the model to produce one of the predefined labels.

We observed that with the previous method, the \textbf{To some extent} label was often under-predicted in favor of the \textbf{Yes} or \textbf{No} labels.
To address this, we introduced an alternative method using thresholds defined separately for each dimension.
We retrained the multi-dimension model (i.e. a single model for the first four tracks) on a subset of the training data and used the remaining examples as a validation set to tune the thresholds.
The training/validation split was 80/20.

Using the fine-tuned model’s predictions on the validation set, we computed the average probability of each class, grouped by the predicted label.
Based on these statistics, we manually defined threshold rules using only the predicted probabilities for the \textbf{Yes} and \textbf{No} labels.
Table~\ref{tab:thresholdsqwen} in Appendix~\ref{sec:appendix_thresholds} shows the threshold values we chose.

\begin{table*}[ht!]
\centering
\scalebox{0.64}{
\begin{tabular}{lcccccccccc}
\hline
& \multicolumn{2}{c}{\textbf{Track 1}} & \multicolumn{2}{c}{\textbf{Track 2}} & \multicolumn{2}{c}{\textbf{Track 3}} & \multicolumn{2}{c}{\textbf{Track 4}} & \multicolumn{2}{c}{\textbf{Track 5}} \\
\textbf{Approach} & \multicolumn{2}{c}{\textbf{Mistake identification}} & \multicolumn{2}{c}{\textbf{Mistake location}} & \multicolumn{2}{c}{\textbf{Guidance}} & \multicolumn{2}{c}{\textbf{Actionability}} & \multicolumn{2}{c}{\textbf{Tutor identity}} \\
 & F1-macro & Accuracy & F1-macro & Accuracy & F1-macro & Accuracy & F1-macro & Accuracy & F1-macro & Accuracy \\
\hline
\multicolumn{9}{l}{\textbf{Preliminary experiments}} \\
\hline
Always Yes & 30.26 & 83.13 & 26.42 & 65.66 & 22.98 & 52.61 & 22.16 & 49.80 & -- & -- \\
Random & 23.95 & 33.73 & 26.94 & 31.93 & 30.02 & 31.33 & 28.99 & 30.72 & 11.75 & 12.05\\
Gemini Flash 2.0 Lite & 50.74 & 76.71 & 48.49 & 62.25 & 46.54 & 55.62 & -- & -- & -- & --\\
Llama 3.1 8B (LoRA) & 74.41 & \textbf{92.37} & 50.68 & 76.71 & 51.83 & \textbf{63.25} & 55.90 & 72.09 & -- & -- \\
\hline
\multicolumn{9}{l}{\textbf{Classical Machine Learning}} \\
\hline
RandomForest + TF-IDF (1-5)grams & 71.46 & 90.76 & 47.24 & 74.10 & 44.08 & 60.64 & 52.00 & 64.26 & 70.25 & 69.68 \\
RandomForest + TF-IDF (1-7)grams & 60.14 & 87.55 & 40.17 & 60.64 & 40.93 & 52.21 & 42.00 & 58.23 & 69.66 & 68.47 \\
RandomForest + TF-IDF (1-8)grams & 71.04 & 90.56 & 46.92 & 74.30 & 44.37 & 60.44 & 50.19 & 62.25 & 68.13 & 66.87 \\
SVC + TF-IDF (1-2)grams& 71.82 & 91.37 & 49.26 & 75.50 & 43.10 & 61.04 & 48.42 & 65.06  & 79.16 & 77.71 \\
SVC + TF-IDF (2-5)grams & 60.47 & 88.96 & 43.74 & 73.96 & 40.68 & 60.64 & 40.92 & 60.04 & 74.76 & 73.69 \\
k-NN ($k=7$) + TF-IDF (1-2)grams & 73.24 & 91.16 & 46.82 & 71.49 & 49.52 & 60.64 & 50.91 & 59.84  & 60.11 & 59.24\\
\hline
\multicolumn{9}{l}{\textbf{Fine-tuning DistilBERT}} \\
\hline
DistilBERT & 58.37 & 90.76 & 50.30 & \textbf{76.91} & 42.24 & 61.24 & 57.01 & 68.47 & 86.51 & 85.94\\
DistilBERT  (thresholds) & 63.04 & 88.35 & \textbf{56.30} & 67.47 & 52.22 & 55.22 & 53.02 & 66.67 & -- & -- \\
BERT Embeddings + XGBoost & -- & -- & -- & -- & -- & -- & -- & -- & \textbf{87.74} & \textbf{87.14} \\ 
\hline
\multicolumn{9}{l}{\textbf{Sentence Embeddings}} \\
\hline

e5 (response) k-NN ($k=9$) & 74.93 & 91.77 & 48.06 & 76.69 & 47.40 & 58.84 & 48.55 & 58.84 & -- & -- \\
e5 (resp+hist) MLP & 72.82 & 90.96 & 54.17 & 75.30 & 52.51 & 60.44 & 56.42 & 65.06 & -- & -- \\
e5 (resp+hist+llm) MLP  & 73.54 & 91.16 & 54.36 & 75.30 & 52.10 & 59.84 & 56.51 & 65.46 & -- & -- \\
e5 (resp) k-NN (balanced) & \textbf{79.16} & \textbf{92.37} & 47.82 & 71.08 & 52.85 & 56.83 & 49.08 & 50.00 & -- & -- \\
\hline
\multicolumn{9}{l}{\textbf{Fine-tuning Qwen}} \\
\hline
Qwen (dimension-specific) & 74.73 & 92.17 & 48.30 & 74.70 & 52.71 & 63.05 & \textbf{61.20} & \textbf{73.29} & -- & -- \\
Qwen (multi-dimension) & 73.04 & 91.37 & 51.96 & 74.50 & 53.26 & 61.45 & 59.57 & 68.47 & -- & -- \\
Qwen (thresholds) & 65.58 & 81.33 & 54.92 & 64.06 & \textbf{54.18} & 55.82 & 55.82 & 58.84 & -- & -- \\
\hline
\end{tabular}
}
\caption{Results for the five tracks over our \texttt{dev} set. In bold, the best results according to each metric for each track.}
\label{tab:dev_results}
\end{table*}

\subsection{Results obtained over the \texttt{dev} set}
\label{sec:results_dev}

We evaluated all these models on the \texttt{dev} set and the results are shown in Table~\ref{tab:dev_results}.

The first observation we want to do is that even when classical ML algorithms did not manage to be the best in any track, they are still competitive.
Some of them even achieved good performances, sometimes getting closer to the best model in the track.
Secondly, we want to highlight that some sentence embeddings approaches performed better than using the fine-tuned Llama3.1 8B that we considered in the preliminary experiments.
Finally, as expected, neural models performed the best.

Another interesting observation is that the \texttt{Llama 3.1 8B} LoRA fine-tuning, a model 16 times larger than Qwen and BERT, did not achieve significantly better results.
In some dimensions, such as providing guidance and actionability, it even performed worse than the fine-tuned Qwen.\\

Overall, the best models in each track were as follows: 
\begin{itemize}
    \setlength\itemsep{0em}
    \item Track 1 (Mistake Identification) -- Sentence Embeddings and k-NN, using the balanced dataset
    \item Track 2 (Mistake Location) -- Fine-tuning DistilBERT with thresholds
    \item Track 3 (Providing Guidance) -- Fine-tuning Qwen using the multi-dimension approach and thresholds
    \item Track 4 (Actionability) -- Fine-tuning Qwen using the multi-dimension apporach
    \item Track 5 (Tutor identification) -- BERT embeddings and XGBoost
\end{itemize}

\section{Final submissions and experimental analysis}
\label{sec:analysis}

After evaluating all the previously described models on our \texttt{dev} set, we chose those which had the best performance, trying to ensure that at least one model of each category (\textit{Classical machine learning}, \textit{Fine-tuning DistilBERT}, \textit{SentenceEmbeddings} and \textit{Fine-tuning Qwen}) was used to predict the test instances in most of the tracks.
The classical ML models were trained from scratch using both our \texttt{train} and \texttt{dev} set, while the neural models were only fine-tuned using the \texttt{train} set.

Table~\ref{tab:submissions} shows the performance of our models in each track, the results we obtained and the resulting ranking position (\#).
To better understand the performance of our systems, we also considered quartiles for each track, and they are included in the table under the ``Q'' column.
Taking a first look at the quartiles, we can see that none of our models was competitive enough to climb the rankings and finish in the first quartile.
However, we want to highlight that in three out of the five tracks (Track1, Track3 and Track4) our models managed to finish in $Q_2$.

\begin{table}[ht!]
\centering
\scalebox{0.69}{
\begin{tabular}{lcccc}
\hline
\textbf{Submission} & \textbf{F1-macro} & \textbf{Accuracy}  & \textbf{\#} & \textbf{Q} \\
\hline
\multicolumn{5}{c}{\textbf{Track 1 - Mistake identification}} \\
\hline
e5 (resp.) k-NN (balanced) & \textbf{65.35} & 84.49 & 56/153 & $Q_2$ \\
Qwen (dimension-specific) & 64.94 & 86.68 & 62/153   & $Q_2$ \\
DistilBERT (thresholds) & 64.30 & 85.20 & 64/153 & $Q_2$\\
SVC + TF-IDF & 59.11 & 84.81 & 104/153 &  $Q_3$\\
e5 (response) k-NN ($k=9$) & 58.39 & 84.36 & 110/153 & $Q_3$ \\
\hline
\multicolumn{5}{c}{\textbf{Track 2 - Mistake location}} \\
\hline
DistilBERT (thresholds) & \textbf{49.58} & 58.63 & 47/86 & $Q_3$ \\
Qwen (multi-dimension) & 49.52 & 70.78 & 49/86 &  $Q_3$ \\
e5 (resp+hist) MLP & 49.40 & 67.36 & 51/86 &  $Q_3$\\
Qwen (thresholds) & 49.13 & 55.20 & 54/86 & $Q_3$ \\
SVC + TF-IDF & 45.85 & 70.39 & 72/86 & $Q_4$ \\
\hline
\multicolumn{5}{c}{\textbf{Track 3 - Providing guidance}} \\
\hline
Qwen (multi-dimension) & \textbf{50.49} & 59.47 & 36/105 & $Q_2$  \\
DistilBERT (thresholds) & 49.19 & 53.85 & 48/105 & $Q_2$  \\
Qwen (thresholds) & 47.53 & 50.36  & 64/105 & $Q_3$ \\
k-NN + TF-IDF & 47.41 & 59.21 & 66/105  &  $Q_3$ \\
e5 (resp+hist) MLP & 47.14 & 57.85 & 71/105  & $Q_3$ \\
\hline
\multicolumn{5}{c}{\textbf{Track 4 - Actionability}} \\
\hline
Qwen (dimension-specific) & \textbf{61.28} & 70.33 & 42/87 & $Q_2$   \\
Qwen (multi-dimension) & 60.54 & 68.00 & 46/87  & $Q_3$ \\
e5 (resp+hist) MLP & 56.37 & 63.22 & 60/87  &  $Q_3$ \\
DistilBERT (thresholds) & 52.61 & 64.12  & 68/87 & $Q_4$ \\
RandomForest + TF-IDF & 51.91 & 62.64 & 70/87 & $Q_4$ \\
\hline
\multicolumn{5}{c}{\textbf{Track 5 - Tutor identification}} \\
\hline
BERT + XGBoost & \textbf{83.85} & 84.74 & 27/54 & $Q_3$  \\
DistilBERT & \textbf{83.85} & 84.74 & 28/54 & $Q_3$   \\
SVC + TF-IDF & 80.44 & 80.22 & 39/54 & $Q_3$  \\
BERT + Educated guess & 68.16 & 68.65 & 42/54 &  $Q_4$ \\
\hline
\end{tabular}
}
\caption{Results for the five tracks over the competition's test data. The ``\#'' column indicates the position the system got in the rankings, and the ``Q'' column indicates the quartile related to that position (splitting in 4 buckets the number of participants in each track).}
\label{tab:submissions}
\end{table}

\begin{table*}[t!]
\centering
\scalebox{0.75}{
\begin{tabular}{|c|c|c|c|c||c|c|}
\hline
\textbf{Track} & \textbf{Rank} / Total & \textbf{Q} & \textbf{$\Delta$ Exact $F_1$}  &  \textbf{$\Delta$ Exact Accuracy} &  \textbf{$\Delta$ Lenient $F_1$} &  \textbf{$\Delta$ Lenient Accuracy}\\ \hline
Track 1 & 23 / 44 & $Q_3$ & $71.81 - 65.35 = 06.46$ & $86.23 - 84.49 = 01.74$ & $89.57 - 83.95 = 05.62 $ & $94.57 - 91.92 = 02.65$ \\ \hline
Track 2 & 21 / 32 & $Q_3$ & $59.83 - 49.59 = 10.24$ & $76.79 - 58.63 = 18.16$ & $83.86 - 72.00 = 11.86 $ & $86.30 - 76.08 = 10.22$ \\ \hline
Track 3 & 17 / 35 & $Q_3$ & $58.34 - 50.49 = 07.85$ & $66.13 - 59.47 = 06.66$ & $77.98 - 70.57 = 07.41 $ & $81.90 - 77.51 = 04.39$ \\ \hline
Track 4 & 17 / 29 & $Q_3$ & $70.85 - 61.29 = 09.56$ & $72.98 - 70.33 = 02.65$ & $85.27 - 82.72 = 02.55 $ & $88.37 - 85.59 = 02.78$ \\ \hline
Track 5 & 12 / 20 & $Q_3$ & $96.98 - 83.85 = 13.13$ & $96.64 - 84.75 = 11.89$ & \textit{N/A} & \textit{N/A} \\ \hline
\end{tabular}
}
  \caption{\label{tab:team_performance_deltas} Performance difference between our best submissions and the winners, for each task. This table was built based on the \textit{team results}, so the total number of submissions for each track is always fewer than those considered in Table~\ref{tab:submissions}.}
\end{table*}

Overall, as when evaluating on the \texttt{dev} set, this time the neural models again achieved the best performance among our models.
Moreover, something interesting to observe is that the fine-tuned Qwen models and the BERT models got similar performance. 
This seems to indicate that the generative capabilities of Qwen are good enough to also work as an emergent classifier.

Most of the models we used are not well-suited for handling long contexts.
This led us to question how essential the \textit{conversation history} truly is for assessing the four evaluation dimensions, or whether the model's response alone is enough to obtain good results.
Therefore, we tested training some models both with and without including the conversation history as input.
In this regard, the most significant experiments were those using sentence embeddings.
These experiments show that nearly every dimension benefits from the inclusion of history, except for the mistake identification dimension, which performs notably better without it (i.e. using only the tutor's response).
More broadly, Qwen-based methods (which incorporate the full conversation history) achieve the best results in providing guidance and actionability, and, in contrast, BERT-based methods (which do not use the conversation history) perform better on mistake location.
This pattern suggests that more subtle dimensions like guidance and actionability benefit more from access to the full conversational context.
Further experimentation is required to validate all these preliminary observations.

Finally, while the DistilBERT with XGBoost approach seemed to have a good performance on our \texttt{dev} set, its final performance (on the test set) was identical to that of the DistilBERT fine-tuning model (without XGBoost).
This was not the only difference we had between our \texttt{dev} set and the \texttt{test} set.
As can be seen by comparing Tables~\ref{tab:dev_results} and ~\ref{tab:submissions}, most methods performed noticeably better on our internal \texttt{dev} set than on the \texttt{test} set.
We believe this performance gap may be due to differences in the class distributions between the two sets.

Furthermore, the experiment using under-sampling to balance the classes showed a significant improvement on the \texttt{test} set, going from being the worst-performing submission to being the best one.
This further highlights the impact of class imbalance on model performance.

\subsection{How far were these lightweight models from winning?}

Finally, to answer our \textit{research question}, we wanted to check how far our lightweight models went in the shared task.
Beyond the ranking positions, we wanted to focus on \textit{how big} (according to the official scores) was the gap between these models and those that settled the state of the art, winning the competition.
In Table~\ref{tab:team_performance_deltas} we show the difference ($\Delta$) --- for each metric --- of our best predictions with the winner team in each track\footnote{
Since the organizers considered the \textit{Exact $F_1$} metric as the main one, we considered as \textit{winning teams} those which got the highest score according to that metric.
Therefore, for all metrics, we calculated the $\Delta$ according to the score achieved by the winning team in that track. This way, even if a different team got a better result according to other metric, we still calculated the $\Delta$ according to the winning team.
}.
As a reference, we also include our team's position in that track and the correspondent \textit{quartile} (this time, based on the number of teams, and not on the number of submitted systems).

Taking a look at the table, we can see that, according to $\Delta$ Exact $F_1$, the closest gap between our performance and the winning team was $06.46$ (in Track 1), while the biggest gap was $13.13$ (in Track 5).
We think this difference in performance is very small considering the restrictions we had.

\section{Conclusions}
\label{sec:conclusions}

In this paper we presented the RETUYT-INCO participation at the 2025 BEA shared task, characterized by our self-imposed restriction of only using models under 1B parameters.
Although our research lab have access to cheap API LLMs and very limited access to run 7B LLMs on clusters, we are conscious that this is not the case for other research labs in the Global South, that usually work in even deeper under-resourced scenarios.
Our self-imposed restriction tries to represent this scenario.

Overall, we used classical machine learning models, BERT-based models, and a QWEN 0.5B LLM.
Despite their (very small) size we finished in mid-ranking positions.
Beyond the results in the rankings, the result we want to highlight is that the gaps in performance we had with the winning teams were between $6.46$ and $13.13$ $F_1\ exact$ points.

We find that gap surprisingly small, taking into account that we did not use LLMs bigger than 1B, nor paid for API access, nor paid for premium cloud computing, nor needed top-tier resources to run our experiments.
Additionally, following the environmental concerns that surround the carbon footprint of state-of-the-art LLMs~\citep{luccioni2023estimating,faizllmcarbon,liu2024green}, we consider this an interesting tradeoff: to sacrifice some performance, in order to have models that do not need extensive training/inference time or power, but that are still competent.
Based on all of the above, we think research on models that run on low-cost GPUs --- or need no GPU at all --- should definitely go on.

\section{Limitations}
\label{sec:limitations}

Throughout the paper we have outlined several limitations we have to run experiments with large models.
These constraints led to our self-imposed restriction of using only neural models with fewer than 1B parameters.
Naturally, our work does not present state-of-the-art results, nor does it intend to. 
Furthermore, we prioritized breadth (i.e. trying many model types) over depth (i.e. optimizing a single approach or architecture extensively). 
While this gives a broader perspective on the diverse possibilities that lightweight models have to offer, it may have limited the performance ceiling of individual models.

Regarding our methodology, we made the decision of splitting the full set into two subsets (\texttt{train} and \texttt{dev}) considering as a priority to keep the conversations and their responses in the same subset.
This decision may have introduced some noise and class imbalance, since we found remarkable differences in the performance of our models over the \texttt{dev} set and the final \texttt{test} set (after submission).
Since the fine-tuned models and the thresholds used were adjusted specifically to our \texttt{dev} set, they may not generalize well to other similar corpora. 

Finally, and related to the previous considerations, we did not systematically perform hyperparameter tuning due to both hardware and time limitations.
Additionally, prior to our final submissions, we only trained (from scratch) the classical ML models on the full set (our \texttt{train} + \texttt{dev} sets).
Since the neural models were our best approaches, searching better hyperparameters and training them with more data could have made the performance gaps a bit smaller.

\section*{Acknowledgments}

This paper has been funded by ANII (Uruguayan Innovation and Research National Agency), Grant No. $FMV\_1\_2023\_1\_176581$.

\bibliography{anthology,custom}

\appendix

\section{Prompts}
\label{sec:appendix_prompts}

This Appendix presents the prompts used for fine-tuning the decoder-only language models, as explained in Section~\ref{sec:ftllm}.\\
 
\noindent\textbf{System prompt}

\begin{verbatim}
You are a critic evaluating a tutor's 
interaction with a student, responsible
for providing a clear and objective single
evaluation based on specific criteria.
Each assessment must accurately reflect the 
absolute performance standards.
\end{verbatim}

\noindent\textbf{User message}

\begin{verbatim}
# Previous Conversation between Tutor and
Student:
{history}

# Scoring Rubric:
{rubric}

# Tutor Response:
{response}
\end{verbatim}

\noindent\textbf{Mistake identification rubric}

\begin{verbatim}
[Has the tutor identified a mistake in a 
student's response?]
1) Yes
2) To some extent
3) No
\end{verbatim}

\noindent\textbf{Mistake location rubric}

\begin{verbatim}
[Does the tutor's response accurately point
to a genuine mistake and its location?]
1) Yes
2) To some extent
3) No
\end{verbatim}

\noindent\textbf{Providing guidance rubric}

\begin{verbatim}
[Does the tutor offer correct and relevant
guidance, such as an explanation, ela-
boration, hint, examples, and so on?]
1) Yes (guidance is correct and relevant
to the mistake)
2) To some extent (guidance is provided but
it is fully or partially incorrect or 
incomplete)
3) No
\end{verbatim}

\noindent\textbf{Actionability rubric}

\begin{verbatim}
[Is it clear from the tutor’s feedback what
the student should do next?]
1) Yes
2) To some extent
3) No
\end{verbatim}

\section{Qwen Thresholds}
\label{sec:appendix_thresholds}

Table~\ref{tab:thresholdsqwen} presents the thresholds used with the Qwen model, as explained in Section~\ref{sec:ftllm}.

\begin{table}[h!]
\centering
\scalebox{0.52}{
\begin{tabular}{|l|c|c|c|}
\hline
\textbf{Dimension} & \textbf{Yes condition} & \textbf{No condition} & \textbf{TSE condition} \\
\hline
Mistake Identification & Yes > 0.90 \& No < 0.05 & Yes < 0.40 \& No > 0.50 & Otherwise \\
Mistake Location       & Yes > 0.75 \& No < 0.15 & Yes < 0.42 \& No > 0.50 & Otherwise \\
Providing Guidance     & Yes > 0.65 \& No < 0.12 & Yes < 0.35 \& No > 0.45 & Otherwise \\
Actionability          & Yes > 0.70 \& No < 0.14 & Yes < 0.25 \& No > 0.65 & Otherwise \\
\hline
\end{tabular}
}
\caption{Threshold-based classification rules for each evaluation dimension using Qwen. TSE = "To some extent".}
\label{tab:thresholdsqwen}
\end{table}

\end{document}